\documentclass[journal]{IEEEtran}
\usepackage{amsmath,amsfonts}
\usepackage{algorithmic}
\usepackage{algorithm}
\usepackage{array}
\usepackage[caption=false,font=normalsize,labelfont=sf,textfont=sf]{subfig}
\usepackage{textcomp}
\usepackage{stfloats}
\usepackage{url}
\usepackage{verbatim}
\usepackage{graphicx}
\usepackage{cite}
\usepackage{adjustbox}
\usepackage{array}
\usepackage{xcolor}
\usepackage{bbding}
\usepackage{pifont}
\newcommand{\cmark}{\ding{51}}%
\newcommand{\xmark}{\ding{55}}%
\usepackage{booktabs}
\usepackage{multirow}
\usepackage{xspace}
\newcommand{\name}[0]{ELLA\xspace}
\usepackage{soul}
\usepackage{tabularx}
\usepackage{orcidlink}

\definecolor{lightblue}{RGB}{221, 231, 245} 
\definecolor{lightyellow}{RGB}{209, 239, 241}
\definecolor{lightgreen}{RGB}{255, 240, 230}
\definecolor{lightred}{RGB}{255,102,102}

\begin{document}

\title{Towards Efficient LLM-aware Heterogeneous\\ Graph Learning}

\author{Wenda Li, Tongya Zheng, Shunyu Liu, Yu Wang, Kaixuan Chen,
Hanyang Yuan, Bingde Hu, 

Zujie Ren, Mingli Song,~\IEEEmembership{Senior Member,~IEEE}, 
Gang Chen,~\IEEEmembership{Senior Member,~IEEE}

\thanks{Wenda Li, Yu Wang, Kaixuan Chen, Hanyang Yuan, Bingde Hu, Mingli Song, and Gang Chen are with State Key Laboratory of Blockchain and Data Security, Zhejiang University (e-mail: lwdup@zju.edu.cn, yu.wang@zju.edu.cn, yuanhanyang@zju.edu.cn, onyhu@zju.edu.cn, brooksong@zju.edu.cn).}
\thanks{Wenda Li is also with Zhejiang Lab.}
\thanks{Zujie Ren is with College of Computer Science and Technology, Zhejiang University and Zhejiang Lab (e-mail:renzju@zju.edu.cn). \emph{(Corresponding author: Zujie Ren.)}}
\thanks{Tongya Zheng is with Zhejiang Provincial Engineering Research Center for Real-Time SmartTech in Urban Security Governance, School of Computer and Computing Science, Hangzhou City University (e-mail: doujiang\_zheng@163.com).}
\thanks{Kaixuan Chen is also with Hangzhou High-Tech Zone (Binjiang) Institute of Blockchain and Data Security, Hangzhou, Zhejiang, China.}
\thanks{Shunyu Liu is with Nanyang Technological University.}
\thanks{Bingde Hu is also with Bangsun Technology.}
}

\markboth{Journal of \LaTeX\ Class Files,~Vol.~XX, No.~X, August~20XX}%
{Shell \MakeLowercase{\textit{et al.}}: A Sample Article Using IEEEtran.cls for IEEE Journals}


\maketitle
\begin{abstract}
Heterogeneous graphs are widely present in real-world complex networks, where the diversity of node and relation types leads to complex and rich semantics.
Efforts for modeling complex relation semantics in heterogeneous graphs are restricted by the limitations of predefined semantic dependencies and the scarcity of supervised signals.
The advanced pre-training and fine-tuning paradigm leverages graph structure to provide rich self-supervised signals, but introduces semantic gaps between tasks.
Large Language Models (LLMs) offer significant potential to address the semantic issues of relations and tasks in heterogeneous graphs through their strong reasoning capabilities in textual modality, but their incorporation into heterogeneous graphs is largely limited by computational complexity.
Therefore, in this paper, we propose an \textbf{E}fficient \textbf{LL}M-\textbf{A}ware (ELLA) framework for heterogeneous graphs, addressing the above issues.
To capture complex relation semantics, we propose an LLM-aware Relation Tokenizer that leverages LLM to encode multi-hop, multi-type relations.
To reduce computational complexity, we further employ a Hop-level Relation Graph Transformer, which help reduces the complexity of LLM-aware relation reasoning from exponential to linear.
To bridge semantic gaps between pre-training and fine-tuning tasks, we introduce the fine-grained task-aware textual Chain-of-Thought (CoT) prompts.
Extensive experiments on four heterogeneous graphs show that our proposed \name outperforms state-of-the-art methods in the performance and efficiency. In particular, \name scales up to 13b-parameter LLMs and achieves up to a 4× speedup compared with existing LLM-based methods.
Our code is publicly available at \url{https://github.com/l-wd/ELLA}.
\end{abstract}


\section{Introduction}
\IEEEPARstart{H}{eterogeneous} graphs are widely used in real-world systems such as recommendation systems~\cite{ni2019justifying,liu2022cross}, academic networks~\cite{ji2021heterogeneous,zhang2015citation}, and knowledge graphs~\cite{wang2021relational,shi2024tgformer}. 
These graphs contain multiple types of nodes and relations, providing rich and complex semantics.
Heterogeneous graph learning is designed to capture these complex semantic relations, including citation relations among papers and authorship relations between papers and authors.

In response to the unique characteristics of heterogeneous graphs, a variety of Heterogeneous Graph Neural Networks (HGNNs) have been developed. Mainstream methods include relation-based~\cite{hong2020attention, hu2020heterogeneous, yang2021interpretable, yu2022heterogeneous}, meta-path-based~\cite{wang2019heterogeneous,fu2020magnn,zhu2020effective,yu2024heterogeneous}, and token-based~\cite{lu2024heterogeneous,mao2023hinormer,kuang2024transformer,liu2021anomaly} approaches. 
Relation-based methods adopt relation-specific aggregation, meta-path-based methods encode handcrafted semantic paths, and token-based methods represent heterogeneous elements as tokens with attention-based aggregation. 
Collectively, these methods attempt to model the complex relation semantics of the textual modality from different perspectives, but they are constrained by predefined semantic dependencies, including meta-paths and inherent heterogeneous relations.
Moreover, they generally require extensive labeled data to effectively learn complex relation semantics, and face severe performance degradation when supervised signals are scarce. 

The pre-training and fine-tuning paradigm~\cite{sun2022gppt} has become a dominant strategy to alleviate the scarcity of supervised signals by leveraging self-supervised signals derived from inherent graph structures~\cite{liu2022graph, wu2021self, yu2024heterogeneous}, but it leads to a substantial gap between pre-training and fine-tuning tasks.
Inspired by the success of the language prompt in textual modality, graph prompt learning~\cite{sun2022gppt,yu2024hgprompt,yu2024generalized} attempts to manipulate downstream data by inserting an additional learnable prompt module, and reformulates fine-tuning tasks as pre-training tasks to bridge the gap between pre-training and fine-tuning tasks.
These prompt modules typically introduce learnable parameters, either as additional node embeddings or task-specific subgraphs, and insert them into the entire graph based on heuristic rules or similarity measures.
However, they mostly operate indiscriminately on the whole graph, lacking fine-grained prompting for each node and relation semantics

Large Language Models (LLMs)~\cite{zhao2023survey,cai2025survey} are particularly promising for capturing complex relation semantics and bridging semantic gaps between tasks, owing to their powerful semantic reasoning and understanding capabilities in the textual modality.
Advanced studies have explored incorporating LLMs into graph learning to enhance the expressiveness and inference capability of graph-based models.
A common strategy is to fine-tune LLMs~\cite{tang2024higpt,zou2023pretraining} to better capture graph structures or to enhance the representational power of graph neural networks (GNNs) through specialized architectures~\cite{huang2024can,wang2024bridging}.
However, the significant computational cost of fine-tuning makes it difficult for these methods to scale even to modestly sized LLMs.
In contrast, prompt-based approaches~\cite{huang2024can,wang2024bridging,zhu2024efficient,gao2025bootstrapping} leverage the reasoning capabilities of LLMs only to capture node semantics, rather than complex relation semantics in heterogeneous graphs, because the exponentially increasing computational complexity caused by neighbor explosion makes such relation modeling difficult.
Therefore, there remain three fundamental challenges unsolved for heterogeneous graph learning. 
\textbf{First, can we effectively harness the reasoning capabilities of LLMs to capture complex relation semantics in heterogeneous graphs?}
\textbf{Second, can we reduce the computational complexity of LLM-based relation reasoning?}
\textbf{Third, can we characterize fine-grained task semantics to seamlessly bridge the semantic gap between pre-training and fine-tuning stages?}

To tackle the above challenges, we propose an \textbf{E}fficient \textbf{LL}Ms-\textbf{A}ware framework (\name) for heterogeneous graph learning. 
Firstly, we design an LLM-aware Relation Tokenizer to capture the complex relation semantics across multi-hop and multi-type in heterogeneous graphs. 
Secondly, the Hop-level Relation Graph Transformer performs hop-wise aggregation of multi-type hidden states derived from LLM-aware relation reasoning, reducing the reasoning complexity from exponential to linear while avoiding semantic confusion.
Finally, to bridge the semantic gap between pre-training and fine-tuning tasks, we introduce fine-grained task-aware textual prompts for each node based on Chain-of-Thought (CoT) reasoning, providing prompts within relation semantics in both stages.
Extensive experiments on multiple heterogeneous graph benchmarks demonstrate that \name, as a prompting-based framework, achieves superior effectiveness and efficiency compared to \textit{state-of-the-art} methods, with an average performance improvement of 3.79\%.
In terms of efficiency, compared with existing LLM-based methods, \name scales up to 13b-parameter LLMs and achieves up to a 4× speedup.

Overall, our contributions are summarized as follows:
\begin{itemize}
    \item 
    We identify three critical challenges in applying LLMs to heterogeneous graph learning: the semantic complexity of relations and tasks when integrating LLMs into heterogeneous graphs, and the excessive computational complexity of LLM-based relation reasoning.
    \item 
    We propose \name, an efficient prompt-based framework for heterogeneous graphs that captures complex relation semantics, performs LLM-based relation reasoning with linear complexity, and uses CoT-based prompts to bridge the gap between pre-training and fine-tuning tasks.
    \item 
    Extensive experiments on diverse heterogeneous graphs show that \name outperforms \textit{state-of-the-art} methods by 3.79\% on average, while efficiently scaling up to 13b-parameter LLMs and achieving up to a 4× speedup over existing LLM-based approaches.
\end{itemize}

\section{Related Work}
\noindent
\subsection{Heterogeneous Graph Neural Networks.}
HGNNs are capable of handling multiple types of nodes and relationships while extracting deep semantic information. Based on different strategies for processing node and relationship information, these models can be categorized into three primary types: relation-based HGNNs~\cite{hong2020attention, hu2020heterogeneous, yang2021interpretable, yu2022heterogeneous}, meta-path-based HGNNs~\cite{yu2014personalized,wang2019heterogeneous,fu2020magnn,zhu2020effective,yu2024heterogeneous}, and token-based heterogeneous graph Transformers~\cite{mao2023hinormer,lu2024heterogeneous,zhu2024hhgt,liu2021anomaly}.
Relation-based HGNNs explicitly model different types of relationships to capture the structural and semantic heterogeneity in heterogeneous graphs. For example, HetSANN~\cite{hong2020attention} learns node embeddings by aggregating neighbor features with varying relational importance.
Meta-path-based HGNNs leverage meta-paths to capture specific semantic dependencies. HAN~\cite{wang2019heterogeneous} employs semantic-level attention mechanisms to compute the importance scores of different meta-paths. 
MAGNN~\cite{fu2020magnn} further aggregates all nodes along a meta-path, considering intermediate nodes as informative and essential for semantic learning.
Token-based heterogeneous graph Transformers generate a token sequence for each target node. PHGT~\cite{lu2024heterogeneous} introduces semantic and global tokens to enhance the learning of semantic relationships and global contextual information.
However, these methods generally require sufficient supervision signals to learn high-quality heterogeneous graph embeddings.

\noindent
\subsection{Graph Pre-training and Fine-tuning.}
Recently, self-supervised learning has emerged as an effective approach to mitigate the challenge of weak supervision signals caused by insufficient labeled data. The pre-training and fine-tuning paradigm has been demonstrated to be effective, where a model is first pre-trained on self-supervised tasks and then fine-tuned with a small amount of labeled data and parameters to adapt to downstream tasks. In homogeneous graphs, both GraphCL~\cite{you2020graph} and SimGRACE~\cite{xia2022simgrace} generate augmented views and minimize the distance between different views. To further bridge the gap between pre-training and fine-tuning, GPPT~\cite{sun2022gppt} introduces additional task-specific and structural markers as graph prompts, thereby aligning node classification tasks with self-supervised link prediction. For heterogeneous graphs, HeCo~\cite{wang2021self} proposes a novel HGNN-based co-contrastive learning mechanism, utilizing network schema and meta-paths as two distinct views to capture both local and high-order structures. SHGP~\cite{yang2022self} introduces a self-supervised learning approach for heterogeneous graphs based on structural clustering. HGPrompt~\cite{yu2024hgprompt} designs dual prompts from both task and heterogeneity perspectives, not only unifying pre-training and fine-tuning tasks but also addressing structural differences between homogeneous and heterogeneous graphs.

\noindent
\subsection{Large Language Models for Graph.}
Given the advanced reasoning capabilities of LLMs, the integration of LLMs and GNNs for complex information extraction has gained significant attention. Based on how LLMs process graph-structured data, existing LLM-enhanced methods can be broadly classified into two distinct approaches.
The first approach~\cite{chen2024graphwiz,ye2023natural,tan2023walklm,tang2024higpt,zou2023pretraining,fang2025uniglm,tang2024graphgpt,he2023harnessing,he2024unigraph,li2024zerog} fine-tunes the LLM to perceive graph information. For example, InstructGLM~\cite{ye2023natural} encodes the geometric structure of graphs using natural language representations and designs instruction-based prompts to fine-tune LLMs. 
WalkLM~\cite{tan2023walklm} generates roughly meaningful textual sequences via random walks and uses them to fine-tune the language model.
HiGPT~\cite{tang2024higpt} enhances the cross-domain generalization of LLMs on heterogeneous graphs using a context-based tokenizer and large-scale instruction tuning to capture semantic relationships.
THLM~\cite{zou2023pretraining} introduces a topology-aware pre-training task that jointly optimizes an LLM and an auxiliary heterogeneous GNN to predict node interactions. 
The second approach~\cite{huang2024can,wang2024bridging,zhu2024efficient,gao2025bootstrapping,liu2023one,he2023explanations,guo2024graphedit,zhao2023graphtext} is a prompt-based method that utilizes natural language descriptions to enable LLMs to reason over the graph, from individual nodes and edges to entire subgraphs. Additionally, it integrates textual or structural tokens to combine the strengths of LLMs with GNNs.
GraphAdapter~\cite{huang2024can} employs a GNN as an adapter and utilizes auto-regressive training on node text for pre-training. 
GraphBridge~\cite{wang2024bridging} enhances fine-grained label comprehension by incorporating contextual textual information, thereby bridging local textual features with global structural dependencies. 
ENGINE~\cite{zhu2024efficient} integrates LLMs and GNNs through a tunable side structure. 
GHGRL~\cite{gao2025bootstrapping} utilizes LLM to automatically summarize and classify different data formats and types for various nodes, ensuring well-aligned node feature representations for GNN.
However, existing LLM-enhanced models mainly focus on homogeneous graphs.
Existing methods for LLM-enhanced heterogeneous graphs suffer from two key limitations, as they often overlook meta-paths that capture high-level semantic information and typically rely on fine-tuning LLMs. In contrast, \name focuses on multi-hop relations while eliminating the need for fine-tuning LLMs.

\section{Preliminaries}

\begin{figure*}[ht]
  \centering
  \includegraphics[width=\linewidth]{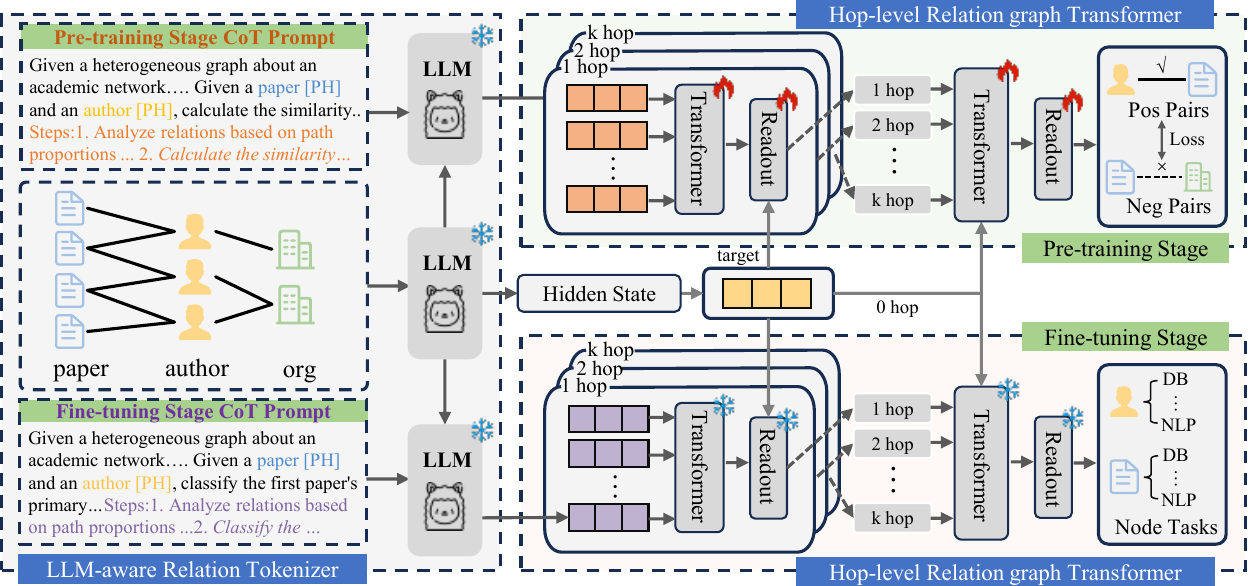}
  \caption{Illustration of our proposed \name. First, we use LLMs to obtain node tokens from node text, and combine \textcolor[HTML]{4F8BCF}{the target node token} with \textcolor[HTML]{FFC247}{the tokens of its $i$-hop neighboring types} to generate relation descriptions and derive relation tokens. Next, we input the relation tokens into a Hop-level Relation Graph Transformer to aggregate rich relation semantics. Finally, we design CoT-based relation prompts for \textcolor[HTML]{EF8843}{pre-training} and \textcolor[HTML]{8266A3}{fine-tuning} to bridge task semantic gaps.}
  \label{graph:main}
\end{figure*}

\noindent
\textbf{Heterogeneous Graph.}
A heterogeneous graph is defined as $\mathcal{G}(\mathcal{V}, \mathcal{E}, \mathcal{X}, \mathcal{A}, \mathcal{R})$, where $\mathcal{V}$ is the set of nodes, and  $\mathcal{E}$ is the set of edges. The sets $\mathcal{A}$ and $\mathcal{R}$ denote the types of nodes and edges, respectively. 
Compared to homogeneous graphs, in heterogeneous graphs, we require $|\mathcal{A}| + |\mathcal{R}| > 2$.
Each node $v \in \mathcal{V}$ has a type $\tau(v) \in \mathcal{A}$  and each edge $e \in \mathcal{E}$ has a type $\phi(e) \in \mathcal{R}$, where $\tau$ is the node type mapping function and $\phi$ is the edge type mapping function, respectively.
$\mathcal{X}$ is the set of textual descriptions of nodes. 

\noindent
\textbf{Meta-path.}
To capture more advanced semantic information, some studies have introduced the concept of meta-path. A meta-path is a specific type of path that connects two nodes, consisting of a sequence of different types of nodes and edges in the graph.
A meta-path of length $l$ is represented as: $A_{1} \stackrel{R_{1}}{\longrightarrow} A_{2} \stackrel{R_{2}}{\longrightarrow} \cdots \stackrel{R_{l}}{\longrightarrow} A_{l+1}$ (abbreviated as $A_1 A_2 \cdots A_{l+1}$ when there is only one unique relation type between $A_i$ and $A_{i+1}$), where $A_i \in \mathcal{A}$ and $R_i \in \mathcal{R}$. It describes a composite relation $R = R_1 \circ R_2 \circ \cdots \circ R_l$ between $A_1$ and $A_{l+1}$.
Note that the meta-path will degenerate into meta meta-relation when $l=1$. Meta-path captures higher-order relationships in heterogeneous graphs by defining path patterns between nodes.

\noindent
\textbf{Multi-Type Node Classification Task.}
The multi-node type classification task refers to the classification task for each node type $\tau(v) \in \mathcal{A}$ in a heterogeneous graph. Unlike traditional single-node type classification, this task requires considering the characteristics and interrelationships of different node types.
In heterogeneous graphs, the representations of heterogeneous neighbors play a crucial role in classifying target nodes. For example, in online academic network recommendation tasks, an author's research field (e.g., Data Mining) is often determined by the venues of his published papers, and vice versa.
However, in real-world scenarios, classifying a single node type often fails to meet practical requirements.
For example, in online academic network tasks, it is not only necessary to classify papers but also to classify the authors of those papers or the organizations they are affiliated with.

\section{Method}
In this section, we introduce the design details of the \name as shown in Figure~\ref{graph:main}.
First, we introduce an LLM-aware Relation Tokenizer to construct semantically rich relation token sequences.
Next, we present the Hop-level Relation Graph Transformer, which reduces the computational complexity of relation reasoning from the perspectives of types and hops, and aggregates relation semantics.
Finally, we present our training strategy, in which fine-grained textual CoT-based prompts are separately designed for pre-training and fine-tuning tasks, leveraging LLMs to bridge the semantic gaps between tasks.

\subsection{LLM-aware Relation Tokenizer}
To adapt to diverse heterogeneous structures, we model multi-hop complex relations rather than relying solely on predefined semantic dependencies including meta-paths or inherent heterogeneous relations, and leverage LLMs to capture complex relation semantics across multi-hop and multi-type, thereby enriching the expressiveness of relation tokens.

Longer contexts increase the inference cost of LLMs~\cite{hooper2025squeezed}, and multi-hop complex relations involving long-text nodes, such as paper abstracts, further exacerbate the context length. Thus, the first step of our approach is to derive informative node tokens, which serve as the foundation for constructing and enriching relation tokens.
Specifically, for nodes rich in textual information, such as paper nodes in academic networks like ACM and DBLP, we leverage their textual content by feeding it into the LLM to extract rich semantic embeddings from its hidden layers. These nodes contain meaningful text attributes, such as titles or abstracts, which provide abundant semantic information to facilitate model learning.
Given a node $s$, the corresponding textual information is $x_s$. The node token $u_s \in \mathbb{R}^d$ is calculated as:
\begin{equation}\label{eq.llm_node}
    \begin{aligned}
        u_s =\operatorname{LLM}(x_s).  
    \end{aligned}
\end{equation}
where $d$ is the hidden dimension.

However, not all node-associated text carries rich semantic information. For example, the textual content of author or organization nodes is often limited to names, which may introduce noise if used directly, since their semantics are primarily derived from neighboring nodes and relation types.
Thus, for a node $t$ without meaningful text, the node token $u_t \in \mathbb{R}^d$ is.
\begin{equation}\label{eq2}
    \begin{aligned}
        u_t =\operatorname{Mean-Pooling}\left(\sum \operatorname{LLM}(x_s), \forall s \in \mathcal{N}_{t}\right).
    \end{aligned}
\end{equation}
where $\mathcal{N}_{t}$ means all neighbors of the target node $t$.
Mean pooling~\cite{ni2022sentence,myers2025lessons,xing2024comparative}, widely used for obtaining semantic representations of long texts, preserves semantics shared across nodes in an LLM-based high-dimensional embedding space, while naturally suppressing node-specific noise through averaging.

After obtaining node tokens, we focus on deriving relation tokens. Unlike existing meta-path-based and relation-based methods that rely on predefined meta-paths and inherent heterogeneous relations, we instead consider paths between arbitrary node pairs within $i$ hop to capture the relation semantics in heterogeneous graphs. 
By leveraging LLMs to extract complex semantics from these paths and encode them as relation tokens, our approach enables flexible and expressive modeling of complex relations in a textual form.
Specifically, given a target node $s$, for its $i$-th hop node $t$ with type $\tau_k$, we collect all possible paths from the target node type to the specified type $\tau_k$ and compute the ratio of paths of the same type. 
The relation token $r^i(s, t) \in \mathbb{R}^d$ is:
\begin{equation}\label{eq.llm_relation}
    \begin{aligned}
        r^i(s, t) =\operatorname{LLM}(\operatorname{RelationPrompt}{(u_s, u_t)}^i).
    \end{aligned}
\end{equation}
where $\operatorname{RelationPrompt}(\cdot)$ is a templated textual prompt designed to describe the $i$-th hop of relation between the two node tokens $u_s$ and $u_t$.

For example, when the hop $i$ is two and both the target node type and the specified type $\tau_k$ are both paper, the $\operatorname{RelationPrompt}{(u_t)}^2$ could be: Given a paper $ u_s $ and another paper $ u_t $, there exists the following paths: $\operatorname{paper} {\xrightarrow{\operatorname{cited}}} \operatorname{paper} {\xrightarrow{\operatorname{cited}}}  \operatorname{paper}$ (proportion of paths: $p_1$), $\operatorname{paper} {\xrightarrow{\operatorname{writed}}} \operatorname{author} {\xrightarrow{\operatorname{write}}} \operatorname{paper}$ (proportion of paths: $p_2$), where $p_1$ and $p_2$ indicate the distribution over 2-hop paths from $u_s$ to $u_t$, with each value representing the proportion of path instances following that specific pattern.

\subsection{Hop-level Relation Graph Transformer}

\begin{table*}
  \caption{Prompts for the $i$-hop relations of the target node. \textcolor[HTML]{4F8BCF}{Color} and \textcolor[HTML]{FFC247}{color} represent the tokens of the two nodes on both sides of a relation, and \textcolor[HTML]{EF8843}{color} denotes the COT-based task prompt.}
  \label{tab:prompts_part}
  \centering
  \renewcommand\arraystretch{1.2} 
  \setlength{\tabcolsep}{1mm}
  \small
  \begin{tabular}{p{0.5cm}p{2.2cm}p{14.3cm}}
    \toprule
    \multicolumn{1}{l}{Hops} & \multicolumn{1}{l}{Source - Target} & \multicolumn{1}{c}{Prompts} \\
    \midrule
    \multirow{9}*{\begin{tabular}[l]{@{}l@{}}one\end{tabular}} & \multirow{4}*{\begin{tabular}[l]{@{}l@{}}paper - author\end{tabular}} & Given a heterogeneous graph about an academic network, there are three types of nodes: paper, author, and organization. The relationships between different nodes include: [author writes paper]... Given a \textcolor[HTML]{4F8BCF}{paper [PH]} and an \textcolor[HTML]{FFC247}{author [PH]}, calculate the similarity based on these paths: paper-author (proportion of paths: $p$). \textcolor[HTML]{EF8843}{Steps: 1. Analyze relations based on path proportions and connection types. 2. Calculate the similarity (0-1) with justification.} \\
    \cmidrule(lr){2-3}
    & \multirow{4}*{\begin{tabular}[l]{@{}l@{}}author - paper\end{tabular}} & Given a heterogeneous graph about an academic network, there are three types of nodes: paper, author, and organization. The relationships between different nodes include: [author writes paper]... Given an \textcolor[HTML]{FFC247}{author [PH]} and a \textcolor[HTML]{4F8BCF}{paper [PH]}, calculate the similarity based on these paths: author-paper (proportion of paths: $p$). \textcolor[HTML]{EF8843}{Steps: 1. Analyze relations based on path proportions and connection types. 2. Calculate the similarity (0-1) with justification.} \\
    \midrule
    \multirow{9}*{\begin{tabular}[l]{@{}l@{}}two\end{tabular}} &\multirow{4}*{\begin{tabular}[l]{@{}l@{}}paper - author\end{tabular}} & Given a heterogeneous graph about an academic network, there are three types of nodes: paper, author, and organization. The relationships between different nodes include: [author writes paper]... Given a \textcolor[HTML]{4F8BCF}{paper [PH]} and an \textcolor[HTML]{FFC247}{author [PH]}, calculate the similarity based on these paths: paper-paper-author (proportion of paths: $p$). \textcolor[HTML]{EF8843}{Steps: 1. Analyze relations based on path proportions and connection types. 2. Calculate the similarity (0-1) with justification.} \\
    \cmidrule(lr){2-3}
    & \multirow{4}*{\begin{tabular}[l]{@{}l@{}}author - paper\end{tabular}} & Given a heterogeneous graph about an academic network, there are three types of nodes: paper, author, and organization. The relationships between different nodes include: [author writes paper]... Given an \textcolor[HTML]{FFC247}{author [PH]} and a \textcolor[HTML]{4F8BCF}{paper [PH]}, calculate the similarity based on these paths: author-paper-paper (proportion of paths: $p$). \textcolor[HTML]{EF8843}{Steps: 1. Analyze relations based on path proportions and connection types. 2. Calculate the similarity (0-1) with justification.}\\
    \bottomrule
  \end{tabular}
\end{table*}
Leveraging the LLM-aware Relation Tokenizer, we generate relation tokens that encapsulate the complex relation semantics between nodes at different hops. 
To reduce the computational complexity caused by neighborhood explosion in relation reasoning, we retain only one type-specific relation token for each target node type in a relation at each hop. For relation tokens of the same relation type at a specific hop, their representations tend to share similar relation semantics. 
For instance, citations from different papers to the target paper typically share similar underlying relation semantics in the LLM-based high-dimensional embedding space, which can be captured by the LLM.

For each target node, we construct a list of generated relation tokens across different hops and relation types.
Specifically, for the target node $s$, the token list at the $i$-th hop is: $U_s^i = \{ u_{s,\tau_1}^i,u_{s,\tau_2}^i,\dots,u_{s,\tau_k}^i\}$ where $\tau_k \in \mathcal{A}$ and $u_{s,\tau_k}^i$ is calculated as follows:
\begin{equation}\label{eq.llm_gt_relation}
    \begin{aligned}
        u_{s,\tau_k}^i &= r^i\left(s, v_{s,\tau_k}^i\right). \\
        v_{s,\tau_k}^i &= \operatorname{Mean-Pooling}\left(\sum u_t, \forall t \in \mathcal{N}_{s, \tau_k}^i\right).
    \end{aligned}
\end{equation}
Here, $\mathcal{N}_{s, \tau_k}^i$ means all nodes of type $\tau_k$ within $i$ hops of the target node $s$. 
We directly compute the single type-specific relation token between the target node $s$ and the pooling node token $v_{s,\tau_k}^i$, which still allows the LLM to infer the underlying relation semantics, rather than computing on every node token found in $\mathcal{N}_{s, \tau_k}^i$. 

Considering that relation tokens at different hops capture neighbor information at different semantic scales~\cite{wang2020multi,chen2024attributed}, we first process the token lists of different types within the same hop.
Specifically, for the $i$-th hop, we first employ a Transformer to compute attention scores across different token types. Subsequently, by utilizing the information from the target node $s$ and a readout function, we obtain the feature token for the $i$-th hop:

\begin{equation} \label{eq:GT_type}
    \begin{aligned}
        \tilde{{U}}_s^{(i,\ell)} &=\mathrm{MHA}\left(\mathrm{LN}\left({U}^{(i,\ell-1)}_{s}\right)\right)+{U}^{(i,\ell-1)}_{s}, \\
        {U}_s^{(i,\ell)} &=\mathrm{FFN}\left(\mathrm{LN}\left( \tilde{{U}}_s^{(i,\ell)}\right)\right)+\tilde{{U}}_s^{(i,\ell)}, \\
        \operatorname{MHA}({U}) &= \operatorname{Concat}\left(\operatorname{head}_0,\operatorname{head}_1,\cdots,\operatorname{head}_B\right){W}_{O}\\
        \operatorname{head}_j &= \operatorname{Softmax}\left( \frac{{U} {W}_j^{Q} ({U} {W}_j^{K})^{\top}}{\sqrt{d_K}}\right){U} {W}_j^{V}.\\
    \end{aligned}
\end{equation}
where $\ell=1, \ldots, L$ implies the $\ell$-th layer of the transformer encoder and the initial input ${U}^{(i,0)}_{s} = U_s^i$, $\operatorname{MHA}(\cdot)$ is Multi-head self-attention, $B$ is the head number, $\mathrm{LN}(\cdot)$ denotes layer normalization and $\mathrm{FFN}(\cdot)$ refers to the feed-forward neural Network. ${W}_j^{Q} \in \mathbb{R}^{d \times d_{K}}, {W}_j^{K} \in \mathbb{R}^{d \times d_{K}}$ and ${W}_j^{V} \in \mathbb{R}^{d \times d_{V}}$ are the projection parameter matrices.

By stacking $L$ layers of transformer, mixed tokens of different relation types can be obtained $\hat{U}_{s}^i = U_s^{(I, L)}$. Then, by calculating the normalized sparse coefficients between the target node $s$ and these mixed tokens, the feature token of the target node $s$ at the $i$-th hop can be further extracted:
\begin{equation}\label{eq.llm_gt_type}
    \begin{aligned}
        h_{s}^i =\underset{\tau_j \in \mathcal{A}} \sum \alpha_{s,\tau_j}^i \cdot \hat{u}_{s,\tau_j}^i, \  \alpha_{s,\tau_j}^i = \frac{\exp(u_s \cdot \hat{u}_{s,\tau_j}^i)}{\underset{\tau_j \in \mathcal{A}} \sum \exp(u_s \cdot \hat{u}_{s,\tau_j}^i)}.
    \end{aligned}
\end{equation}

Based on the target node token and the tokens between different relation types obtained at each hop, we can derive the final sequence of $K+1$ tokens
$H_s = \{h_s^{0},h_s^{1},h_s^{2},\dots,h_s^{K}\}$ where $h_s^0$ is $u_s$, as the $0$-hop corresponds to the node $s$ itself. 
\begin{equation} \label{eq:GT_hop}
    \begin{aligned}
        \tilde{{H}}_s^{(\ell)} &=\mathrm{MHA}\left(\mathrm{LN}\left({H}^{(\ell-1)}_{s}\right)\right)+{H}^{(\ell-1)}_{s}, \\
        {H}_s^{(\ell)} &=\mathrm{FFN}\left(\mathrm{LN}\left( \tilde{{H}}_s^{(\ell)}\right)\right)+\tilde{{H}}_s^{(\ell)}.
    \end{aligned}
\end{equation}

By stacking $L$ layers of transformer, we derive the token for all hops $ \hat{H}_{s} = H_s^{(L)}$.
Here, $\hat{h}_s^0$ is the token of the target node itself, and then the attention coefficients between it and the tokens within $K$ hops are computed, further obtaining the final token for the target node.
\begin{equation}\label{eq.final_output}
    \begin{aligned}
        z_s = \hat{h}_s^0 + \sum_{j = 0}^{K} \gamma_h^i \cdot \hat{h}_s^j, \gamma_h^i = \frac{\exp((\hat{h}_s^0 \Vert \hat{h}_s^i)W^{\top})}{\sum_{j = 1}^{K} \exp((\hat{h}_s^0 \Vert \hat{h}_s^j)W^{\top})}.
    \end{aligned}
\end{equation}
where $W^{\top}$ is the learning weight, $K$ is the hop numbers.

\noindent
\subsection{LLM-aware Pre-training and Fine-tuning}
For a heterogeneous graph, we first employ the LLM-aware Relation Tokenizer to obtain relation tokens. Then, we utilize the Hop-level Relation Graph Transformer to compute the final token for the target node. 
Furthermore, to alleviate the scarcity of supervision signals, \name adopts the pre-training and fine-tuning paradigm with self-supervised signals and designs CoT-based, fine-grained task-aware textual semantic prompts to bridge the gap between pre-training and fine-tuning tasks.

\noindent
\subsubsection{Pre-training Stage}
We mainly use the structural information of the heterogeneous graph as the pre-training objective. Considering the diversity of node and relation types in a heterogeneous graph, we independently sample a set of existing edges for each relation type as positive samples, and generate corresponding negative samples by constructing edges of the same relation type that do not exist in the graph. Then, we adopt a contrastive learning strategy to compute the loss between positive and negative samples, thereby enhancing the model’s representation capability as follows:
\begin{equation}\label{eq.sim_loss}
    \begin{aligned}
        \operatorname{sim}(s,t) &= \operatorname{Sigmoid}(z_s {W_{\tau(s)}} \cdot z_t {W_{\tau(t)}}), \\
        \mathcal{L} &= - \underset{(s,t) \in \text{Pos}} \sum \operatorname{log}(\operatorname{sim}(s,t))  \\ &  + \underset{(s',t') \in \text{Neg}} \sum \operatorname{log}((1 - \operatorname{sim}(s',t'))).
    \end{aligned}
\end{equation}
where $s$ and $t$ are the two nodes connected by the sample edge.

For the pre-training task, we construct CoT-based textual prompts and inject them into $\operatorname{RelationPrompt}$ to provide fine-grained prompts for each target node and multi-hop relation semantics. These prompts are designed to guide the model in reasoning over relation semantics and are tailored for link prediction tasks. Specifically, each prompt follows a two-step reasoning process:
\textbf{Steps: 1. Analyze relations based on path proportions and connection types. 2. Calculate the similarity (0–1) with justification.} 
This design serves two purposes: on the one hand, the first step captures the complex semantics of specific path structures in the heterogeneous graph; on the other hand, the second step estimates similarity, aligns with the objective of link prediction, providing a learning signal grounded in both structure and semantics.
The textual prompt strategy not only preserves semantic space consistency between the prompts and the original textual attributes but also facilitates fine-grained analysis and understanding of each target node.
More examples of textual semantic prompts are provided in Table~\ref{tab:prompts_part}, with additional instances included in Appendix A.

\noindent
\subsubsection{Fine-tuning Stage}
During fine-tuning, the discrepancy between fine-tuning and pre-training tasks renders the pre-trained prompt ineffective. Thus, we design another CoT-based, fine-grained task-aware textual semantic prompt for fine-tuning, as detailed in Appendix A.
Moreover, in this stage, \name only introduces a classification head for the fine-tuning task while keeping other model parameters frozen. For specific node type classification, the model is optimized by minimizing the following cross-entropy loss:
\begin{equation}\label{eq.node_class}
    \begin{aligned}
    \mathcal{L} = - \frac{1}{N} \underset{s \in \mathcal{V}_{\text{label}}} \sum \sum_{c=1}^{C} y_{s,c} \log \hat{y}_{s,c}.
    \end{aligned}
\end{equation}
Here, $N$ is the number of samples, and $C$ is the number of classes. $ \mathcal{V}_{\text{label}} $ denotes the labeled nodes of the specified type. $y_{s,c}$ is the ground-truth label of sample $s$ for class $c$, while $y_s$ represents the class probability obtained from a classifier (e.g., a single-layer neural network) with $z_s$ as input.

\noindent
\subsection{Complexity Analysis}
During model execution, the overall time complexity of \name can be divided into two parts. 
First, in the LLM-aware Relation Tokenizer, the LLM hidden states are computed by performing inference for each node across all relation types and hop levels, resulting in a complexity of $O(n(EK+1)\mathcal{M})$,
where $K$ is the number of hops, $n$ is the number of nodes, $E$ is the number of edge types in the graph, and $O(\mathcal{M})$ denotes the inference time required by the LLM to generate the token of a single node. 
Second, in the Hop-level Relation Graph Transformer, attention is computed over different relation types with a complexity of $O(nE^2Kd)$, and over hop-level embeddings with a complexity of $O(n(K+1)^2d)$, $d$ is the hidden dimension.
Therefore, the overall time complexity of \name is $O\big(n[E^2Kd + (K+1)^2d + (EK+1)\mathcal{M}]\big)$.

\noindent
\section{Experiments}
\noindent
\subsection{Experimental Setup}
\noindent
\subsubsection{Datasets}
Our experiments are conducted on four widely used heterogeneous graph datasets. The statistical information of these datasets is summarized in Table~\ref{tab:datasets}.

\textit{IMDB~\cite{fu2020magnn}.} This dataset represents a movie network consisting of three types of nodes: movies, directors, and actors. Each movie, director, and actor is assigned to one of three categories: action, drama, or comedy based on the primary genre of their films. The IMDB dataset is used for both \textit{node classification} and \textit{link prediction} tasks in our experiments.

\textit{ACM\footnote{\url{www.aminer.cn/citation}\label{web}}.} The ACM dataset is an academic network composed of two types of nodes: authors and papers. Each node is categorized into one of three research domains based on publication venues: database, wireless communication, and data mining. This dataset is used for \textit{node classification}.

\textit{DBLP\textsuperscript{~\ref {web}}.} The DBLP dataset, another academic network, contains four types of nodes: authors, papers, organizations, and keywords. Authors and papers are classified into four research areas according to publication venues: database, machine learning, web and information retrieval, and natural language processing. The DBLP dataset is also used for \textit{node classification}.

\textit{AMMI~\cite{ni2019justifying}.} The AMMI (Amazon Musical Instruments) dataset is a product review network including two types of nodes: users and items. Edges represent review interactions, capturing ratings and user preferences for modeling behavioral and recommendation patterns. This dataset is used for \textit{link prediction}.

For the node classification task, we conduct multi-type node classification, i.e., classifying at least two types of nodes in each dataset. To ensure consistency across experiments, all datasets adopt the same training and validation set partitioning strategy: 100 nodes per class are selected for training and validation, while the remaining nodes are used as the test set. Classification performance is evaluated using Micro-F1 (\%) and Macro-F1 (\%) metrics.
For the link prediction task, 80 percent of the edges are selected as positive samples. These are then split into training, validation, and test sets in a ratio of 8:1:1. For each part, negative sampled edges are generated in twice the quantity of positive ones using the same ratio to enable a comprehensive evaluation of link prediction. Performance is measured using AUC (\%) and AP (\%) as evaluation metrics.
To ensure the robustness of the experimental results, all experiments are independently repeated three times, and the average results along with the corresponding standard deviations are reported.

\begin{table*}[]
    \caption{Experimental results for multi-type node classification. The classification performance is evaluated using Micro-F1 (\%) and Macro-F1 (\%) metrics, abbreviated as Mi-F1 and Ma-F1 , respectively. The best results are highlighted in bold. ``OOM'' stands for ``Out of Memory'', indicating that memory usage has exceeded the allocated limit. ``OOT'' stands for ``out of time''.  ``$^\ast$'' indicates significantly better performance than the baseline at the 0.05 level, validated by two-sample t-tests over three independent runs. The best results are highlighted in \textbf{bold}. The second-best results are \underline{underlined}. }
    \label{tab:main_node}
    \renewcommand\arraystretch{0.8} 
    \setlength{\tabcolsep}{0.8mm}
    \adjustbox{width=\textwidth}{
        \begin{tabular}{cllllllllllllll}
            \toprule
            \multirow{3}{*}{Method} & \multicolumn{6}{c}{IMDB} & \multicolumn{4}{c}{ACM} & \multicolumn{4}{c}{DBLP} \\
            \cmidrule(lr){2-7}\cmidrule(lr){8-11}\cmidrule(lr){12-15}   
            & \multicolumn{2}{c}{director}  & \multicolumn{2}{c}{movie}  & \multicolumn{2}{c}{actor} & \multicolumn{2}{c}{paper} & \multicolumn{2}{c}{author} & \multicolumn{2}{c}{author} & \multicolumn{2}{c}{paper}  \\
            \cmidrule(lr){2-7}\cmidrule(lr){8-11}\cmidrule(lr){12-15}   
            & \multicolumn{1}{c}{Mi-F1} & \multicolumn{1}{c}{Ma-F1} & \multicolumn{1}{c}{Mi-F1} & \multicolumn{1}{c}{Ma-F1} & \multicolumn{1}{c}{Mi-F1} & \multicolumn{1}{c}{Ma-F1} & \multicolumn{1}{c}{Mi-F1} & \multicolumn{1}{c}{Ma-F1} & \multicolumn{1}{c}{Mi-F1} & \multicolumn{1}{c}{Ma-F1} & \multicolumn{1}{c}{Mi-F1} & \multicolumn{1}{c}{Ma-F1} & \multicolumn{1}{c}{Mi-F1} & \multicolumn{1}{c}{Ma-F1} \\
            \midrule
            GCN             & 45.5{\tiny $\pm$0.6} & 45.3{\tiny $\pm$0.5} & 52.1{\tiny $\pm$1.1} & 52.0{\tiny $\pm$1.1} & 50.3{\tiny $\pm$0.3} & 50.3{\tiny $\pm$0.3} & 88.6{\tiny $\pm$0.1} & 88.2{\tiny $\pm$0.1} & 78.2{\tiny $\pm$0.1} & 78.1{\tiny $\pm$0.1} & 58.1{\tiny $\pm$0.4} & 53.2{\tiny $\pm$0.4} & 71.2{\tiny $\pm$0.3} & 61.3{\tiny $\pm$0.2}         \\
            GAT             & 47.9{\tiny $\pm$0.8} & 47.7{\tiny $\pm$0.8} & 51.1{\tiny $\pm$1.3} & 50.6{\tiny $\pm$1.0} & 53.4{\tiny $\pm$1.2} & 53.3{\tiny $\pm$1.2} & \underline{89.0{\tiny $\pm$0.1}} & \underline{88.6{\tiny $\pm$0.1}} & 83.0{\tiny $\pm$0.2} & 83.0{\tiny $\pm$0.2} & 54.8{\tiny $\pm$1.8} & 50.9{\tiny $\pm$1.3} & 70.7{\tiny $\pm$1.9} & 60.9{\tiny $\pm$1.3}         \\
            NAGphormer      & 47.2{\tiny $\pm$0.8} & 46.8{\tiny $\pm$1.0} & 54.6{\tiny $\pm$0.3} & 53.9{\tiny $\pm$0.4} & 54.1{\tiny $\pm$0.2} & 54.1{\tiny $\pm$0.3} & 89.0{\tiny $\pm$0.2} & 88.6{\tiny $\pm$0.2} & 85.8{\tiny $\pm$0.3} & 85.9{\tiny $\pm$0.3} & 59.9{\tiny $\pm$0.1} & 54.1{\tiny $\pm$0.1} & 69.6{\tiny $\pm$2.9} & 60.5{\tiny $\pm$2.6}         \\
            \midrule
            HetSANN         & 48.0{\tiny $\pm$0.6} & 47.9{\tiny $\pm$0.7} & 53.8{\tiny $\pm$0.6} & 53.6{\tiny $\pm$0.6} & 54.0{\tiny $\pm$0.6} & 54.1{\tiny $\pm$0.6} & 82.5{\tiny $\pm$0.3} & 81.8{\tiny $\pm$0.2} & 77.4{\tiny $\pm$0.2} & 77.1{\tiny $\pm$0.2} & 62.9{\tiny $\pm$0.5} & 56.2{\tiny $\pm$0.3} & 66.5{\tiny $\pm$0.2} & 55.5{\tiny $\pm$0.1}         \\
            R-HGNN          & 46.4{\tiny $\pm$1.0} & 46.4{\tiny $\pm$1.0} & 53.1{\tiny $\pm$0.5} & 52.5{\tiny $\pm$1.4} & 53.6{\tiny $\pm$0.3} & 53.8{\tiny $\pm$0.3} & 86.6{\tiny $\pm$2.0} & 86.1{\tiny $\pm$1.9} & 86.6{\tiny $\pm$2.0} & 86.1{\tiny $\pm$1.9} & 61.4{\tiny $\pm$1.7} & 55.2{\tiny $\pm$1.3} & 63.9{\tiny $\pm$0.9} & 54.3{\tiny $\pm$1.1}         \\
            HINormer        & 32.9{\tiny $\pm$3.3} & 29.7{\tiny $\pm$2.0} & 34.7{\tiny $\pm$1.3} & 34.1{\tiny $\pm$1.8} & 33.7{\tiny $\pm$2.7} & 30.9{\tiny $\pm$1.4} & 50.0{\tiny $\pm$3.2} & 46.3{\tiny $\pm$1.8} & 31.8{\tiny $\pm$8.2} & 28.6{\tiny $\pm$7.2} & 22.2{\tiny $\pm$5.3} & 18.7{\tiny $\pm$3.4} & 27.6{\tiny $\pm$1.0} & 22.3{\tiny $\pm$0.5}         \\
            \midrule
            SimGRACE        & 45.8{\tiny $\pm$0.2} & 44.9{\tiny $\pm$0.1} & 52.3{\tiny $\pm$0.8} & 51.2{\tiny $\pm$0.8} & 51.8{\tiny $\pm$1.5} & 51.6{\tiny $\pm$1.4} & 83.9{\tiny $\pm$1.1} & 83.7{\tiny $\pm$1.0} & 75.8{\tiny $\pm$0.8} & 75.7{\tiny $\pm$0.8} & 51.8{\tiny $\pm$3.6} & 47.6{\tiny $\pm$2.9} & 65.3{\tiny $\pm$1.6} & 56.5{\tiny $\pm$1.0}         \\
            GraphCL         & 40.4{\tiny $\pm$1.3} & 40.1{\tiny $\pm$1.2} & 46.9{\tiny $\pm$1.1} & 46.6{\tiny $\pm$0.9} & 49.1{\tiny $\pm$0.1} & 48.5{\tiny $\pm$0.5} & 86.7{\tiny $\pm$1.3} & 86.3{\tiny $\pm$1.4} & 80.7{\tiny $\pm$0.1} & 80.5{\tiny $\pm$0.1} & 53.0{\tiny $\pm$0.4} & 47.5{\tiny $\pm$0.3} & 63.2{\tiny $\pm$1.5} & 52.8{\tiny $\pm$1.0}         \\
            \midrule
            HeCo            & 36.0{\tiny $\pm$2.3} & 36.1{\tiny $\pm$1.5} & 52.1{\tiny $\pm$0.9} & 51.5{\tiny $\pm$0.9} & 37.2{\tiny $\pm$0.6} & 37.2{\tiny $\pm$0.4} & 59.4{\tiny $\pm$0.4} & 59.2{\tiny $\pm$0.3} & 67.0{\tiny $\pm$0.3} & 66.0{\tiny $\pm$0.3} & OOM                  & OOM                  & OOM                  & OOM                          \\
            SHGP            & 35.4{\tiny $\pm$3.9} & 34.9{\tiny $\pm$1.9} & 42.4{\tiny $\pm$0.9} & 43.7{\tiny $\pm$1.0} & 34.5{\tiny $\pm$3.2} & 36.7{\tiny $\pm$0.7} & 61.8{\tiny $\pm$0.9} & 60.9{\tiny $\pm$0.9} & 62.9{\tiny $\pm$0.9} & 62.3{\tiny $\pm$0.8} & 33.4{\tiny $\pm$3.0} & 30.1{\tiny $\pm$1.4} & 34.3{\tiny $\pm$3.5} & 27.9{\tiny $\pm$1.4}         \\
            \midrule
            GPPT            & 38.3{\tiny $\pm$1.2} & 36.5{\tiny $\pm$0.7} & 37.7{\tiny $\pm$2.1} & 37.0{\tiny $\pm$2.0} & 33.6{\tiny $\pm$0.0} & 30.4{\tiny $\pm$0.0} & 80.9{\tiny $\pm$0.2} & 80.3{\tiny $\pm$0.1} & 71.5{\tiny $\pm$0.0} & 71.4{\tiny $\pm$0.0} & 19.8{\tiny $\pm$0.0} & 16.1{\tiny $\pm$0.0} & 61.2{\tiny $\pm$0.1} & 42.0{\tiny $\pm$0.1}         \\
            HGPROMPT        & 37.0{\tiny $\pm$1.1} & 34.9{\tiny $\pm$0.0} & 36.2{\tiny $\pm$1.0} & 35.2{\tiny $\pm$0.8} & 34.8{\tiny $\pm$0.7} & 36.4{\tiny $\pm$0.2} & 63.2{\tiny $\pm$0.3} & 61.7{\tiny $\pm$0.2} & 34.9{\tiny $\pm$0.4} & 32.2{\tiny $\pm$1.1} & 39.7{\tiny $\pm$0.1} & 35.2{\tiny $\pm$0.3} & 57.1{\tiny $\pm$0.1} & 47.8{\tiny $\pm$0.0}         \\
            \midrule
            GraphAdapter    & \underline{67.3{\tiny $\pm$0.5}} & 65.8{\tiny $\pm$0.6} & \underline{77.1{\tiny $\pm$0.7}} & \underline{76.6{\tiny $\pm$0.6}} & 45.6{\tiny $\pm$0.8} & 45.3{\tiny $\pm$0.7} & 88.7{\tiny $\pm$0.2} & 88.5{\tiny $\pm$0.2} & 44.0{\tiny $\pm$1.5} & 41.9{\tiny $\pm$2.0} & OOM                  & OOM                  & OOM                  & OOM                          \\
            GraphBridge     & 47.7{\tiny $\pm$1.5} & 46.5{\tiny $\pm$1.3} & 47.2{\tiny $\pm$0.6} & 47.3{\tiny $\pm$0.6} & 46.7{\tiny $\pm$2.2} & 45.6{\tiny $\pm$2.1} & 88.3{\tiny $\pm$0.6} & 88.2{\tiny $\pm$0.5} & 85.6{\tiny $\pm$0.5} & 85.7{\tiny $\pm$0.5} & 62.0{\tiny $\pm$1.7} & 53.4{\tiny $\pm$1.0} & 69.5{\tiny $\pm$1.4} & 59.0{\tiny $\pm$1.0}         \\
            ENGINE          & 49.7{\tiny $\pm$1.5} & 48.4{\tiny $\pm$1.1} & 51.2{\tiny $\pm$1.5} & 50.9{\tiny $\pm$1.7} & 49.9{\tiny $\pm$2.2} & 49.2{\tiny $\pm$2.1} & 87.8{\tiny $\pm$0.5} & 87.6{\tiny $\pm$0.4} & 87.1{\tiny $\pm$0.4} & 87.2{\tiny $\pm$0.4} & OOM                  & OOM                  & OOM                  & OOM                          \\
            THLM            & 57.2{\tiny $\pm$0.6} & 57.1{\tiny $\pm$0.4} & 59.4{\tiny $\pm$0.4} & 59.4{\tiny $\pm$0.6} & 61.2{\tiny $\pm$0.1} & \underline{60.9{\tiny $\pm$0.2}} & 88.0{\tiny $\pm$0.1} & 87.7{\tiny $\pm$0.1} & \underline{87.5{\tiny $\pm$0.4}} & \underline{87.4{\tiny $\pm$0.5}} & \underline{79.5{\tiny $\pm$0.5}} & \underline{71.0{\tiny $\pm$1.0}} & \underline{82.0{\tiny $\pm$0.4}} & \underline{70.6{\tiny $\pm$0.5}}         \\
            HiGPT           & 36.4{\tiny $\pm$2.2}         & 32.1{\tiny $\pm$1.4}         & 31.4{\tiny $\pm$2.6}         & 26.5{\tiny $\pm$2.1}         & 34.8{\tiny $\pm$3.3}         & 31.9{\tiny $\pm$2.7}         & OOT                          & OOT                          & OOT                          & OOT                          & OOT                          & OOT                          & OOT                          & OOT                          \\
            GHGRL           & 66.7{\tiny $\pm$1.1}         & \underline{66.0{\tiny $\pm$1.1}}         & 66.4{\tiny $\pm$0.6}         & 65.8{\tiny $\pm$0.5}         & \underline{61.4{\tiny $\pm$0.5}}         & 60.6{\tiny $\pm$0.7}         & OOT                          & OOT                          & OOT                          & OOT                          & OOT                          & OOT                          & OOT                          & OOT                          \\

            \midrule
            ELLA            & \textbf{79.3$^\ast${\tiny $\pm$0.1}} & \textbf{78.4$^\ast${\tiny $\pm$0.2}} & \textbf{80.5$^\ast${\tiny $\pm$0.5}} & \textbf{80.3$^\ast${\tiny $\pm$0.3}} & \textbf{73.3$^\ast${\tiny $\pm$0.1}} & \textbf{73.3$^\ast${\tiny $\pm$0.3}} & \textbf{89.7$^\ast${\tiny $\pm$0.1}} & \textbf{89.4$^\ast${\tiny $\pm$0.1}} & \textbf{88.5$^\ast${\tiny $\pm$0.2}} & \textbf{88.6$^\ast${\tiny $\pm$0.4}} & \textbf{81.5$^\ast${\tiny $\pm$0.1}} & \textbf{73.1$^\ast${\tiny $\pm$0.6}} & \textbf{83.4$^\ast${\tiny $\pm$0.4}} & \textbf{72.4$^\ast${\tiny $\pm$0.3}}        \\ 
            \bottomrule
        \end{tabular}
    }
\end{table*}

\begin{table}[]
    \caption{Heterogeneous Graph Datasets and Statistics. ``\#ATL" denotes the average token length of the node texts.}
    \label{tab:datasets}
    \small
    \centering    
    \adjustbox{width=\linewidth}{
        \begin{tabular}{cllll}
            \toprule
            Dataset               & \#Nodes   & \#Edges         & \#Class     & \#ATL     \\ 
            \midrule
            \multirow{3}{*}{IMDB} & movie : 4278    & movie - actor : 12828   & movie : 3  & 27.9          \\
                                  & actor : 5257    & movie - director : 4278    & director : 3&     -           \\
                                  & director : 2081 &                &  actor : 3        &          -          \\ 
            \midrule
            \multirow{2}{*}{ACM}  & paper : 13866   & paper - paper : 42794 & author : 3 & 228.3 \\
                                  & author : 20359 & author - paper : 43107 & paper : 3 &    -   \\
            \midrule
            \multirow{4}{*}{DBLP} & paper : 37074   & paper - paper : 19583   & author : 4  & 235.5         \\
                                  & author : 71842  & author - paper : 126673  & paper : 4   &    -              \\
                                  & org : 19611     & paper - keyword : 56735  &   -       &     -             \\
                                  & keyword :34673  & author - org : 76890  &    -      &         -          \\
            \midrule
            \multirow{2}{*}{AMMI} & user : 10611    & user - item : 231312  & -   &          -          \\
                                  & item : 27530    &                &        &  189.3                 \\
            \bottomrule
        \end{tabular}
    }
\end{table}
\noindent
\subsubsection{Baselines.}
To comprehensively evaluate the proposed \name against state-of-the-art methods, we consider multiple categories of advanced approaches. (1) We include classical homogeneous graph neural networks such as GCN~\cite{kipf2016semi}, GAT~\cite{velivckovic2017graph}, and NAGphormer~\cite{chen2022nagphormer}. Homogeneous graph pre-training and fine-tuning frameworks, such as SimGRACE~\cite{xia2022simgrace}, GraphCL~\cite{you2020graph}, and GPPT~\cite{sun2022gppt}. (2) We incorporate classical HGNNs, including HetSANN~\cite{hong2020attention} and R-HGNN~\cite{yu2022heterogeneous}. Heterogeneous graph pre-training frameworks, including HeCo~\cite{wang2021self}, SHGP~\cite{yang2022self}, HINormer~\cite{mao2023hinormer} and HGPROMPT~\cite{yu2024hgprompt}. (3) We examine joint training frameworks for LLMs and GNNs under homogeneous graphs, including GraphAdapter~\cite{huang2024can}, GraphBridge~\cite{wang2024bridging}, and ENGINE~\cite{zhu2024efficient}. For heterogeneous graphs, the representative frameworks are THLM~\cite{zou2023pretraining}, GHGRL~\cite{gao2025bootstrapping}, and HiGPT~\cite{tang2024higpt}.

\noindent
\subsubsection{Implementation Details}
In the implementation of \name, the hidden dimension of the model is uniformly set to 128. The number of multi-head attention layers for processing relation type tokens is set to two, while the number of multi-head attention layers for handling hop tokens is set to three. During the pretraining stage, the learning rate is fixed at 1e-4. During the fine-tuning stage, the learning rate is selected by searching within the range of 0.01, 0.001, and 1e-4 to determine the optimal value. Both training stages adopt early stopping, with the patience parameter set to 30.

\noindent
\subsubsection{Running Environment}
All experiments were conducted on a Linux server equipped with an NVIDIA A40 GPU and 96 GB of memory. The software environment includes Python 3.9 and CUDA 11.6. 

\noindent
\subsection{Overall Performance}
\noindent
\subsubsection{Node Classification}
The results in Table~\ref{tab:main_node} show that our proposed \name method achieves exceptional performance across three heterogeneous datasets.
First, compared to traditional homogeneous and heterogeneous graph neural network methods, \name exhibits a significant performance improvement. This validates the effectiveness of leveraging LLMs to encode node text and different types of relations in heterogeneous graphs through tokenization. Extracting semantic information between relations via LLMs significantly enhances model performance.
Furthermore, compared to joint training schemes that integrate homogeneous and heterogeneous graphs with LLMs, our method also achieves substantial improvements. This further confirms that \name effectively handles the relation complexity of heterogeneous graphs, capturing high-order semantic information from diverse node and relation types. 
Since our training set is substantially larger than that used for HiGPT in its few-shot training setting, HiGPT suffers from training instability that leads to performance degradation. This is consistent with the findings reported in Table 2 of the HiGPT paper, where increasing the amount of training data results in unstable training and lower performance.
\noindent
\subsubsection{Link Prediction}
Table~\ref{tab:main_link} presents the link prediction performance of different methods on two heterogeneous graph datasets, IMDB and AMMI. It can be observed that \name outperforms both traditional GNN approaches and existing methods that incorporate LLMs on the link prediction task.
Compared with other traditional methods and LLM-based approaches, \name better captures the complex structures and heterogeneous semantics of heterogeneous graphs by introducing multi-hop relation modeling. Leveraging the deep semantic reasoning capabilities of LLMs, it captures complex relation semantics and generates more semantically rich token representations, leading to significantly improved performance in both AUC and AP.

\begin{table}[]
    \caption{Experimental results for link prediction. The prediction performance is evaluated using AUC(\%) and AP(\%) metrics. The best results are highlighted in bold. ``–'' indicates experimental failure.}
    \label{tab:main_link}
    \centering
    \adjustbox{width=\linewidth}{
        \begin{tabular}{lcccc}
        \toprule
        \multicolumn{1}{c}{\multirow{2}{*}{Method}} & \multicolumn{2}{c}{IMDB}                            & \multicolumn{2}{c}{AMMI}             \\
        \cmidrule(lr){2-5}
                     & \multicolumn{1}{c}{AUC(\%)}                    & \multicolumn{1}{c}{AP(\%)}                      & \multicolumn{1}{c}{AUC(\%)}                     & \multicolumn{1}{c}{AP(\%)}                       \\
                     
        \midrule
        GCN          & 89.3{\tiny $\pm$ 0.5} & 73.7{\tiny $\pm$ 0.9} & 86.7{\tiny $\pm$ 0.7} & 81.1{\tiny $\pm$ 0.5} \\
        GAT          & 81.8{\tiny $\pm$ 1.5} & 79.9{\tiny $\pm$ 1.5} & 54.3{\tiny $\pm$ 5.0} & 45.3{\tiny $\pm$ 7.4} \\
        NAGphormer   & 90.6{\tiny $\pm$ 1.6} & 73.4{\tiny $\pm$ 3.6} & 95.7{\tiny $\pm$ 0.1} & 89.9{\tiny $\pm$ 0.1} \\
        \midrule
        HetSANN      & 95.4{\tiny $\pm$ 0.4} & 95.3{\tiny $\pm$ 0.3} & \underline{96.1{\tiny $\pm$ 0.2}} & 95.2{\tiny $\pm$ 0.1} \\
        R-HGNN       & \underline{97.3{\tiny $\pm$ 0.4}} & \underline{97.1{\tiny $\pm$ 0.7}} & 95.7{\tiny $\pm$ 0.8} & \underline{95.8{\tiny $\pm$ 0.7}} \\
        HINormer     & 80.6{\tiny $\pm$ 0.1} & 70.2{\tiny $\pm$ 0.8} & 78.1{\tiny $\pm$ 0.8} & 67.1{\tiny $\pm$ 0.1} \\
        SHGP         & 51.1{\tiny $\pm$ 3.1} & 34.3{\tiny $\pm$ 2.6} & 50.1{\tiny $\pm$ 0.2} & 33.4{\tiny $\pm$ 0.1} \\
        HGPROMPT     & 49.9{\tiny $\pm$ 1.2} & 33.2{\tiny $\pm$ 0.8} & -                     & -                     \\
        \midrule
        GraphAdapter & 80.4{\tiny $\pm$ 0.7} & 66.7{\tiny $\pm$ 1.7} & 73.6{\tiny $\pm$ 0.2} & 65.3{\tiny $\pm$ 0.2} \\
        GraphBridge  & 79.9{\tiny $\pm$ 8.9} & 75.2{\tiny $\pm$ 2.0} & -                     & -                     \\
        ENGINE       & 96.1{\tiny $\pm$ 0.2} & 92.4{\tiny $\pm$ 0.4} & -                     & -                     \\
        THLM         & 93.7{\tiny $\pm$ 0.7} & 93.4{\tiny $\pm$ 0.9} & 95.3{\tiny $\pm$ 0.5} & 95.1{\tiny $\pm$ 0.7} \\
        GHGRL        & 49.5{\tiny $\pm$ 0.1} & 32.8{\tiny $\pm$ 0.2} & -                     & -                     \\
        \midrule
        ELLA         & \textbf{97.5{\tiny $\pm$ 0.5}} & \textbf{97.3{\tiny $\pm$ 0.4}} & \textbf{96.5$^\ast${\tiny $\pm$ 0.1}} & \textbf{96.1{\tiny $\pm$ 0.2}} \\
        \bottomrule
    \end{tabular}
    }
\end{table}

\begin{table}[]
    \caption{Efficiency comparison of combining different LLMs with graph-based methods. ``FT" indicates whether fine-tuning of LLMs is required. ``Mem" is the used memory, measured in GB. ``Time" represents the time taken to train one epoch of the model, measured in milliseconds (ms).}
    \label{tab:efficient}
    \centering
    \begin{tabular}{lllcr}
        \toprule
        Method                       & FT & Mem   & LLMs                & Time (ms)    \\
        \midrule
        GraphAdapter                 & \xmark  & 29.44 & LLaMA-2-13b            & 84.94  \\
        \midrule
        \multirow{2}{*}{GraphBridge} & \cmark  & 21.70 & LLaMA-2-13b + BERT     & 127.10 \\
                                     & \cmark  & OOM   & LLaMA-2-7b/13b & -       \\
        \midrule
        \multirow{2}{*}{ENGINE}      & \xmark  & 19.52 & LLaMA-2-7b             & 810.71 \\
                                     & \xmark  & OOM   & LLaMA-2-7b/13b            & -       \\
        \midrule
        \multirow{2}{*}{THLM}        & \cmark  & 0.62  & BERT                & 419.94 \\
                                     & \cmark  & OOM   & LLaMA-2-7b/13b            & -       \\
        \midrule
        \name                        & \xmark  & 6.44  & LLaMA-2-13b            & 45.66  \\
        \bottomrule
    \end{tabular}
\end{table}

\begin{figure}[h]
  \centering
  \includegraphics[width=\linewidth]{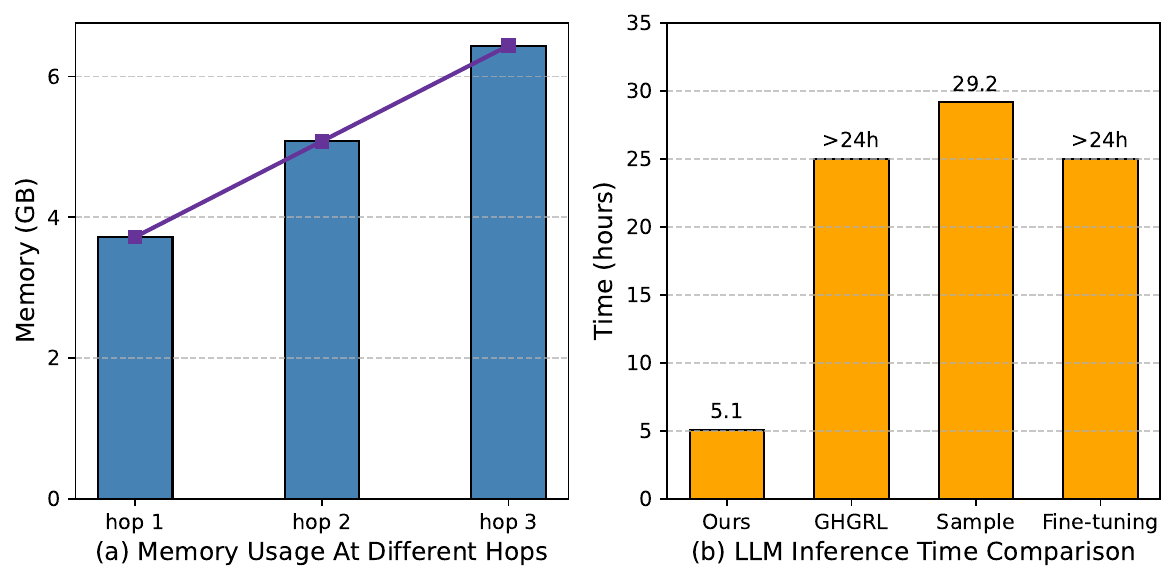}
  \caption{Efficiency comparison of memory usage and LLM inference time.}
  \label{fig:time_mem}
\end{figure}

\subsection{Efficiency Studies}
In this section, we evaluate the efficiency of methods leveraging LLMs. When jointly training LLMs with GNNs, the large parameter size of LLMs often becomes the efficiency bottleneck, while GNNs are relatively lightweight. Therefore, we compare different methods from three aspects: (1) whether the LLM is fine-tuned, which incurs a significant computational cost, (2) the memory consumption after introducing the LLM, and (3) the training time overhead caused by the LLM integration. We conduct experiments on the ACM dataset using an NVIDIA A40 GPU. 

As shown in Table \ref{tab:efficient}, (1) among the methods involving LLM fine-tuning, only the BERT-based model can be successfully fine-tuned due to its moderate size. 
In contrast, larger LLMs with extensive parameters, such as LLaMA-2-7b, fail to fine-tune because of GPU memory limitations, resulting in unsuccessful scalability.
(2) Among the methods without LLM fine-tuning, \name achieves the lowest memory consumption.
This advantage stems from our framework design, which avoids saving a large number of token-level outputs from the LLM. When using relatively large LLMs, our method avoids memory overflow issues, demonstrating good scalability and efficiency. 
Moreover, to assess the impact of the hop number $K$ on memory usage, Figure~\ref{fig:time_mem} (a) reports the memory consumption under different $K$ values. We observe an approximately linear increase in memory with $K$ for \name, which further indicates that \name is largely unaffected by the neighborhood explosion problem. 
The mechanism of saving only lightweight hidden states and linear memory growth in relation reasoning enables \name to scale up to larger LLMs and graphs.
(3) In terms of training time, compared with other LLM-based methods, \name significantly reduces the time per epoch, further validating its efficiency and strong scalability in scenarios of joint training with LLMs.
Figure~\ref{fig:time_mem} (b) reports the LLM-related runtime across different strategies. GHGRL leverages LLM-generated responses to enhance GNN model performance. Sample denotes the inference time spent by the LLM when, at each hop, no more than ten meta-paths are sampled. Fine-tuning refers to the time required to fine-tune the LLM. 
As shown, \name incurs the lowest runtime compared to other LLM-based strategies, achieving a 4x speedup, primarily because it learns from pooled embeddings, which significantly reduces the number and cost of LLM inferences.

\begin{table}[]
    \caption{Ablation studies of \name key components. ``\#v1" replaces the relation tokens at each hop with the corresponding node tokens. ``\#v2" inputs all relation tokens into the Transformer jointly, without distinguishing hops. ``\#v3" retains only the relation prompt for LLMs. ``\#v4" removes the path-specific description from the relation prompt.}
    \label{tab:ablation}
    \small
    \centering
    \setlength{\tabcolsep}{0.9mm}
    \begin{tabular}{cccccccc}
        \toprule
        Dataset               &      Type                 & Metric  & \#v1 & \#v2 & \#v3 & \#v4 & ELLA   \\
        \midrule
        \multirow{4}{*}{IMDB} & \multirow{2}{*}{director} & Mi-F1(\%) & 77.90       & 47.50        & 78.79  & 78.86       & 79.25 \\
                              &                           & Ma-F1(\%) & 77.06       & 46.05        & 77.83  & 77.85       & 78.36 \\
                              \cmidrule(lr){2-8}
                              & \multirow{2}{*}{movie}    & Mi-F1(\%) & 78.83       & 53.28        & 79.72  & 79.80       & 80.51 \\
                              &                           & Ma-F1(\%) & 78.65       & 53.17        & 79.51  & 79.67       & 80.28 \\
        \midrule
        \multirow{4}{*}{ACM}  & \multirow{2}{*}{paper}    & Mi-F1(\%) & 88.24       & 87.68        & 88.12  & 88.43       & 89.72 \\
                              &                           & Ma-F1(\%) & 88.00       & 87.37        & 87.74  & 88.17       & 89.44 \\
                              \cmidrule(lr){2-8}
                              & \multirow{2}{*}{author}   & Mi-F1(\%) & 86.18       & 88.21        & 87.14  & 87.43       & 88.55 \\
                              &                           & Ma-F1(\%) & 86.41       & 88.31        & 87.23  & 87.49       & 88.62 \\
        \bottomrule
        
    \end{tabular}
\end{table}

\begin{figure}[h]
  \centering
  \includegraphics[width=\linewidth]{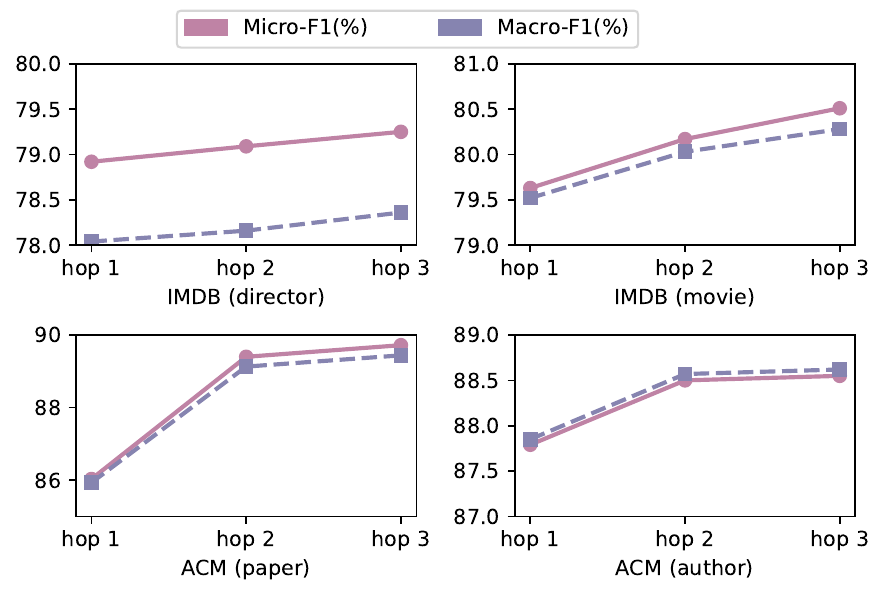}
  \caption{Analysis study on the number of hops.}
  \label{graph:hops}
\end{figure}

\subsection{Ablation Studies} 
To evaluate the effectiveness of the \name model, we conduct ablation studies by individually removing different components to assess the impact of each key component on the model’s performance. The results are summarized in Table~\ref{tab:ablation}.
Here are the ablated variants and the key conclusions:

\noindent
\subsubsection{Effect of LLM-aware Relation Tokenizer} To validate the necessity of the LLM-aware Relation Tokenizer, we construct a variant ``\#v1" by replacing the relation tokens at each hop with node tokens. We observe that using only node tokens leads to a drop in performance, mainly because the model cannot capture the diverse and complex relations in heterogeneous graphs. In contrast, the relation tokens in \name combine node and relation information and leverage LLMs to generate richer semantic representations, which help the model better capture important complex relation semantics.

\noindent
\subsubsection{Effect of Hop-level Relation Graph Transformer} To evaluate the effectiveness of the Hop-level Relation Graph Transformer in our model, we construct a variant ``\#v2" by feeding all relation tokens jointly into the transformer without distinguishing between different hop levels. The results indicate a significant performance degradation when the distinction between different hop levels is removed. This is attributed to the substantial semantic scale differences among relation tokens at different hops. The design of \name effectively captures and distinguishes these differences, leading to improved model performance.

\begin{figure}[!t]
  \centering
  \includegraphics[width=\linewidth]{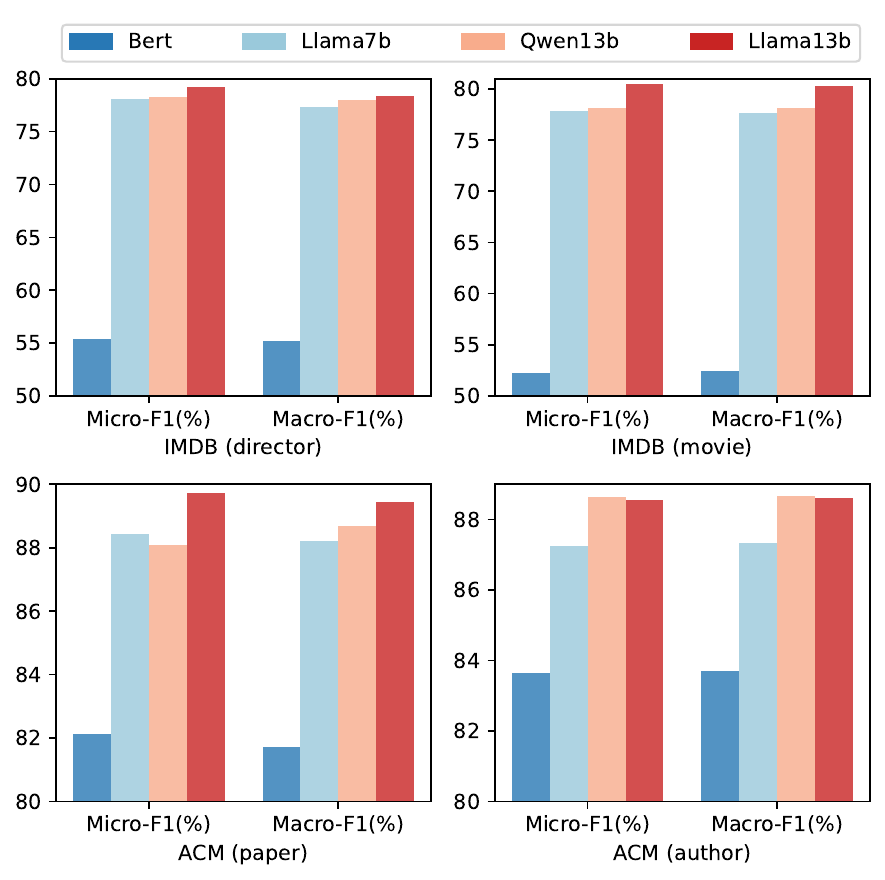}
  \caption{Analysis study on LLMs choices.}
  \label{graph:LLMs}
\end{figure}

\begin{figure}[t]
  \centering
  \includegraphics[width=\linewidth]{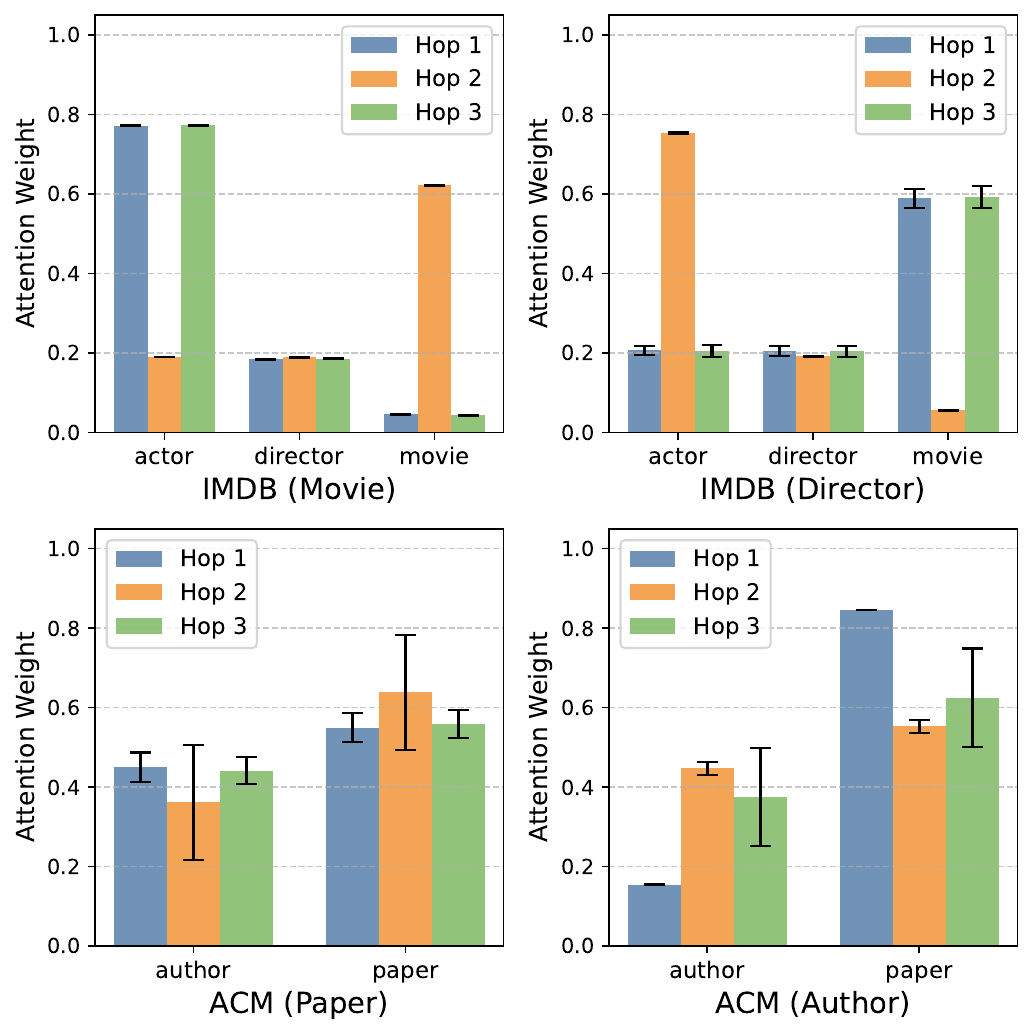}
  \caption{Visualization of attention weight by relation type across hop 1–3 for two central node types in IMDB (movie, director) and ACM (paper, author). Bars show means (error bars: std). The x-axis displays only the target node types of each relation, where the source is the central node type.}
  \label{vis:ralation}
\end{figure}

\begin{figure*}[h]
  \centering
  \includegraphics[width=\linewidth]{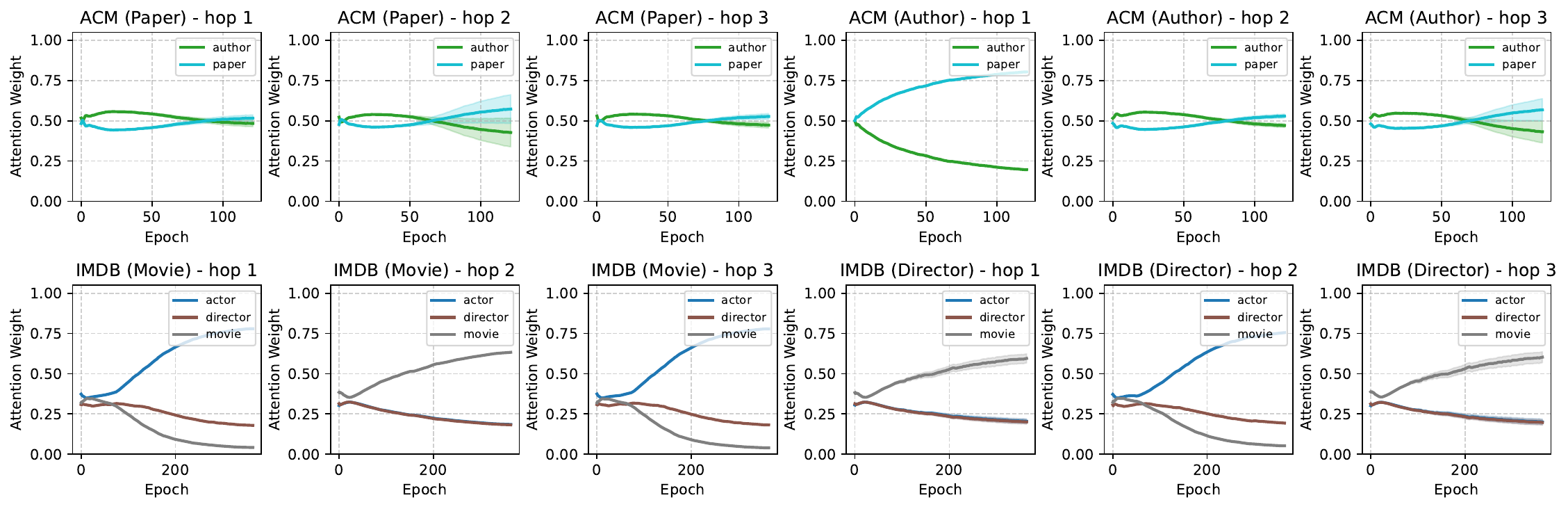}
  \caption{Visualization of attention weights across relation types during training. The solid lines denote the mean attention values, and the shaded areas indicate the standard deviations.}
  \label{vis:ralation_train}
\end{figure*}

\begin{figure}[h]
  \centering
  \includegraphics[width=\linewidth]{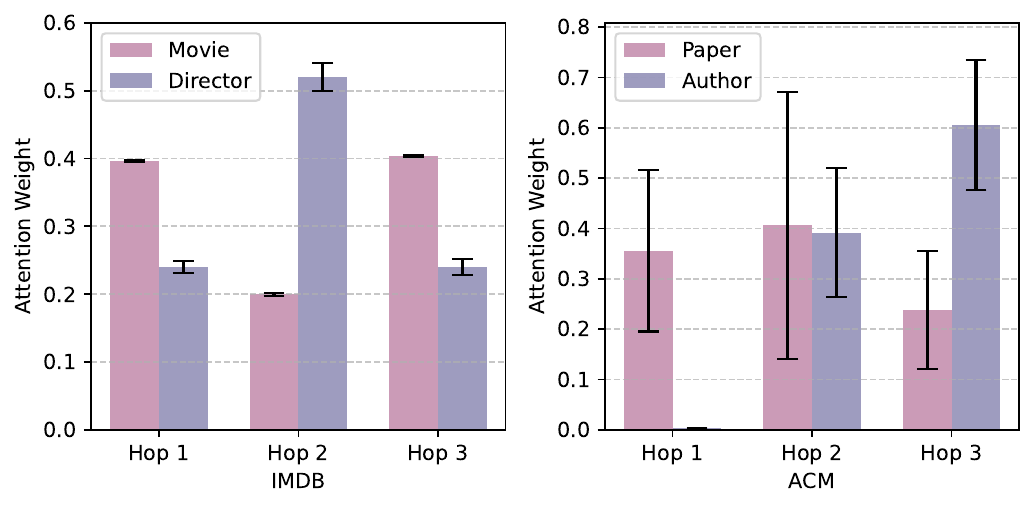}
  \caption{Visualization of attention weights across hops for two central node types in IMDB (movie, director) and ACM (paper, author).}
  \label{vis:hops}
\end{figure}

\begin{figure}[h]
  \centering
  \includegraphics[width=\linewidth]{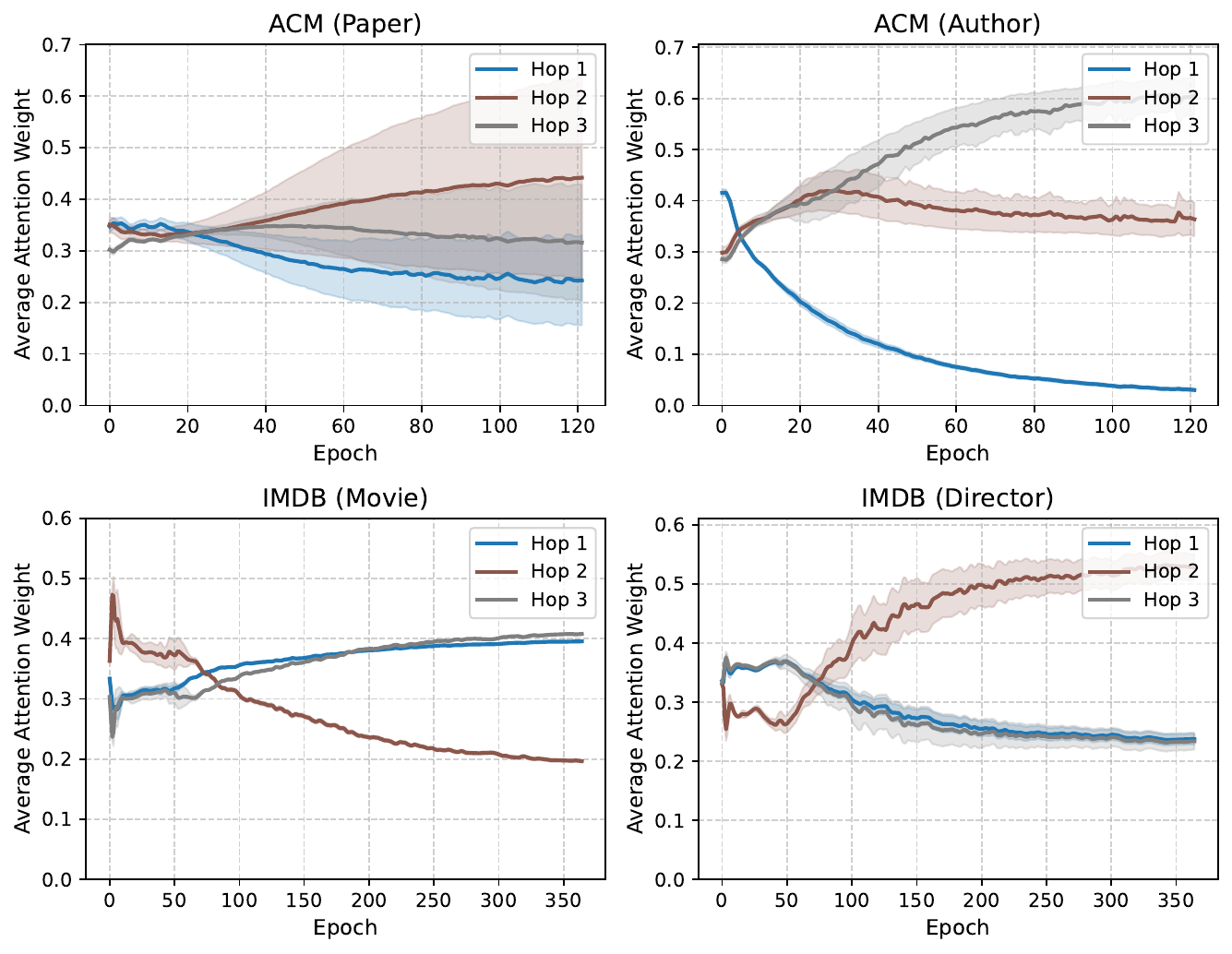}
  \caption{Visualization of attention weights across hops during training. The solid lines denote the mean attention values, and the shaded areas indicate the standard deviations.}
  \label{vis:hops_train}
\end{figure}

\noindent
\subsubsection{Effect of COT-based Task Prompts} We evaluate the necessity of using COT-based task prompts to reduce the task gap. By retaining only the relation-based descriptions in the prompts provided to LLMs, we obtain a variant named ``\#v3". We observe that across various datasets, \name consistently outperforms the variant. This suggests that COT-based task prompts play a crucial role in guiding LLMs to generate task-aligned reasoning tokens, which in turn provide the model with relevant task-specific information and lead to improved performance.

\noindent
\subsubsection{Effect of Path-specific Description} 
To further assess the effectiveness of relation prompts, we constructed a variant ``\#v4" that removes the meta-path descriptions from the prompt. The results show a performance degradation, confirming that explicit meta-path descriptions are not merely auxiliary but encode semantics that guide relation reasoning. This indicates that providing explicit meta-path priors serves as an important inductive bias, enabling the LLM to better model heterogeneous relation patterns and improving its overall inference performance.

\begin{table*}[h]
    \vspace{-0.00in}
    \centering
    \caption{Visualization of our \name's relation-tokenizer response for the author–author relation at hop 3, focused on the first author node in the ACM dataset.}
    \vspace{-0.1in}
    \label{tab:case_1}
    \small
    \begin{tabularx}{\textwidth}{X}
        \toprule
        \textbf{Question For Link Prediction:} Given an author [PH] and an author [PH], predict the likelihood of a connection based on these paths: author-paper-paper-author (Proportion of paths: 0.00 ). Steps:1.Analyze relations based on path proportions and connection types. 2. Predict connection likelihood (0-1) with justification.\\
        Action \\
        \midrule
        (LLM Response): 0.00, which means there is no connection between these two authors. \\
        \midrule
        \textbf{Question For Node Classification:} Given an author [PH] and an author [PH], classify the first paper's primary research field (Database, Wireless Communication, or Data Mining) based on these paths: author-paper-paper-author (Proportion of paths: 0.00 ). Steps:1.Analyze relations based on path proportions and connection types. 2.Classify the first paper's primary research field (Database, Wireless Communication, or Data Mining) with justification.\\
        Action \\
        \midrule
        (LLM Response): Data Mining. The primary research field of the first author is: Data Mining.\\
        \bottomrule
    \end{tabularx}
    \vspace{-0.1in}
\end{table*}

\subsection{Parameter Sensitivity Analysis}
\subsubsection{Analysis on number of hops} 
The length of the input relation token sequence determines the number of hops accessible to the target node, thereby influencing the size of its receptive field. To examine the effect of varying hops, we conducted experiments with hop ranges from 1 to 3, as illustrated in Figure~\ref{graph:hops}. The results demonstrate that the optimal number of hops for maximizing performance varies across datasets. Furthermore, even within the same dataset, the ideal hop count differs for different types of node classification tasks. This variation arises from the distinct neighborhood structures inherent to different datasets and node types.

\subsubsection{Analysis on LLMs choices} 
Figure~\ref{graph:LLMs} presents an analysis of the impact of different LLMs on performance, namely BERT~\cite{kenton2019bert}, LLaMA-2-7b~\cite{touvron2023llama}, Qwen-13b~\cite{chiang2023vicuna}, and LLaMA-2-13b~\cite{touvron2023llama}. Compared to the weaker reasoning model, such as BERT, the other three models demonstrate significant performance improvements across all tasks. This improvement can be attributed to the stronger reasoning capabilities of LLMs, which are particularly evident in complex text understanding tasks. As the model size increases from 7b to 13b, performance further improves. Notably, although Qwen-13b and LLaMA-13b are of the same 13b scale, the former performs slightly worse, which may be due to differences in pretraining corpora or fine-tuning tasks.

\subsection{Visualization Analysis}
In this section, we investigate how the Hop-level Relation Graph Transformer learns relation tokens across different relation types and hop levels.
\subsubsection{Attention Visualization Across Relation Types} 
Figure~\ref{vis:ralation} reports the distribution of attention over relation types across hop levels. On IMDB, movie nodes allocate greater attention to movie-actor at hop 1 and hop 3, possibly because the actor set has a larger and more information-rich dataset compared to the director set. At hop 2, attention concentrates on movie-movie, reflecting the absence of valid meta-paths from movie to actor or director at this hop and consequently making the corresponding relation tokens uninformative. A similar pattern is observed for director-centered subgraphs. On ACM, both author and paper nodes emphasize author-paper and paper-paper, consistent with the graph’s structural asymmetry: paper nodes exhibit broad connectivity to all node types, whereas author nodes primarily connect to paper. Figure~\ref{vis:ralation_train} further indicates that, during training, the model converges to these hop-specific relation preferences, suggesting that it reliably internalizes the underlying structural regularities.

\subsubsection{Attention Visualization Across Hops} 
Figure~\ref{vis:hops} illustrates the distribution of attention across hop levels. On IMDB, movie nodes allocate more attention to hops 1 and 3, while director nodes focus primarily on hop 2. This disparity reflects the available relation types at each depth: at hop 2 from a movie node, the only relation is movie-movie, which provides limited additional information, whereas the hop 2 neighbor relations centered around directors include both director-director and director-actor relations. Similarly, on ACM, author nodes allocate minimal attention to hop 1, as the only 1-hop relation is author-paper. Figure~\ref{vis:hops_train} tracks the attention changes across different hops during training, showing that the model consistently converges to the same hop-specific attention patterns, in line with the above analysis.

\subsection{Case Study}
We further conduct a case study to illustrate the responses of the LLM in \name to relation prompts specifically designed for link prediction and node classification tasks, thereby exploring the role of relation prompts in different tasks.
From Table~\ref{tab:case_1}, we observe that the LLM produces meaningful responses for both link prediction and node classification. 
Moreover, it effectively identifies and leverages the path-related content and distribution described in the relation prompt, reasoning based on the information provided by CoT. For link prediction, the LLM produces a connection probability along with an explanation, reflecting its relation reasoning capability. Conversely, in node classification, the absence of direct relations leads the model to rely mainly on the central node’s token for prediction.

\section{Conclusion}
This paper proposes an efficient LLM-aware framework named \name to improve semantic understanding in heterogeneous graph learningan efficient LLM-aware framework to improve semantic understanding in heterogeneous graph learning 
It leverages an LLM-aware Relation Tokenizer to generate reasoning results from LLMs over complex relations. 
The Hop-level Relation Graph Transformer reduces the computational complexity for LLM-based relation reasoning from exponential to linear.
Furthermore, CoT-based fine-grained textual semantic prompts are employed to bridge the gap between pre-training and fine-tuning tasks.
Experimental results demonstrate that \name outperforms existing methods in both performance and efficiency.

\section{Limitations and Discussions}\label{sec:limit-discu}
In this paper, we propose an efficient LLM-aware framework to improve semantic understanding in heterogeneous graph learning, but its current design is mainly optimized for intra-domain scenarios. In practice, real-world heterogeneous graphs exhibit substantial diversity in both node and relation types, resulting in diverse relation semantics across domains. For example, academic graphs typically follow an author–paper schema, whereas e-commerce graphs are often organized around a user–item pattern. Such discrepancies make direct transfer across domains challenging. A promising direction for future work is to investigate joint learning and transferability mechanisms for heterogeneous graphs from different domains, so as to enhance the model’s ability to adapt to diverse graph structures and to advance heterogeneous graph learning from domain-specific settings toward more universal, cross-domain applications.

\bibliographystyle{IEEEtran}
\bibliography{IEEEabrv}

\end{document}